# A GENERALIZED HYBRID REAL-CODED QUANTUM EVOLUTIONARY ALGORITHM BASED ON PARTICLE SWARM THEORY WITH ARITHMETIC CROSSOVER

Md. Amjad Hossain, Md. Kowsar Hossain and M.M.A Hashem

Department of Computer Science and Engineering
Khulna University of Engineering & Technology (KUET), Khulna, 9203, Bangladesh
amjad_kuet@yahoo.com, auvikuet@yahoo.com and mma_hashem@hotmail.com

## ABSTRACT

*This paper proposes a generalized Hybrid Real-coded Quantum Evolutionary Algorithm (HRCQEA) for optimizing complex functions as well as combinatorial optimization. The main idea of HRCQEA is to devise a new technique for mutation and crossover operators. Using the evolutionary equation of PSO a Single-Multiple gene Mutation (SMM) is designed and the concept of Arithmetic Crossover (AC) is used in the new Crossover operator. In HRCQEA, each triploid chromosome represents a particle and the position of the particle is updated using SMM and Quantum Rotation Gate (QRG), which can make the balance between exploration and exploitation. Crossover is employed to expand the search space, Hill Climbing Selection (HCS) and elitism help to accelerate the convergence speed. Simulation results on Knapsack Problem and five benchmark complex functions with high dimension show that HRCQEA performs better in terms of ability to discover the global optimum and convergence speed.*

## 1. INTRODUCTION

Complex optimization functions characterized by high dimension, non-linear, no convexity, non-differential, a large number of local optima and so on are often used model of many engineering problems. The typical deterministic optimization methods are not suitable for solving complex functions because of its high sensitivity to initial guess and local convergence. Evolutionary Algorithms (EAs) such as Genetic Algorithms (GA), inspired by biological evolution use techniques such as mutation, selection and crossover for optimizing complex functions and have been widely used in many areas for solving practical problems to replace the deterministic optimization methods. But unfortunately, EAs often has the problems of premature convergence and to trap into local optimum. Complex numerical functions as well as many real world problems such as Face detection [1], [2], Disk allocation method [3] etc. are not effectively and efficiently solvable by EAs. A novel evolutionary algorithm known as Quantum Evolutionary Algorithm (QEA) is developed to overcome the shortcomings of EAs. It is characterized on the basis of the concept and the principles of quantum computing such as qubits and superposition of states [4] – [7]. Although it has a better characteristic of diversity in the population than EA, it has been observed that QEA is suitable for combinatorial optimization problems such as knapsack problem, but it traps into local optima during solving multi-peaks complex optimization functions [8], [9].

Particle Swarm Optimization (PSO) is a stochastic optimization technique modeled on swarm intelligence. This technique is developed based on fish schooling, bird flocking etc. PSO has no evolution operators such as crossover and mutation like GA. In PSO, the potential solutions, called particles, fly through the problem space by following the current optimum particles. PSO is easy to implement than QEA and there are few parameters to adjust and has faster convergence speed [10], [11]. PSO has been extended to optimize functions [12], control system [13], parameters optimization of fuzzy system [14], etc. To optimize the complex function effectively and efficiently, real-coded quantum evolutionary algorithm (RCQEA) is proposed in [15]. In RCQEA, real-coded triploid chromosome is used to keep the diversity of the solution instead of qubit chromosome which is used in QEA. Complementary Double Mutation Operator (CDMO) and Quantum Rotation Gate (QRG) are used to update chromosomes which make the balance between the exploration and exploitation. Discrete Crossover (DC) is employed to expand the search space and Hill-climbing selection (HCS) helps to accelerate the convergence speed [15].

Using the advantages of PSO and RCQEA, this paper proposes a Generalized Hybrid Real-coded Quantum Evolutionary Algorithm (HRCQEA) to solve complex numerical problems as well as combinatorial optimization problems. HRCQEA maintains a population of real-coded triploid chromosomes. Each chromosome is considered as a particle which is a potential solution. The Single- Multiple gene Mutation (SMM) is designed using the evolutionary equation of PSO. SMM and Quantum Rotation Gate (QRG) are used to update the position of particle which can treat the balance between exploration and exploitation more effectively and efficiently. A new crossover mechanism based on Arithmetic crossover (AC) is used to expand the search space. Hill Climbing Selection (HCS) and elitism are used to accelerate the convergence speed. Simulation results on five benchmark problems and knapsack problem show that HRCQEA performs better than QEA, PSO, and RCQEA in terms of convergence speed, search capability, and adaptability with dimensions.

The paper is organized as follows. Section 2 includes the related work and Section 3 describes some recent evolutionary techniques such as QEA, PSO, and RCQEA. In Section 4, the mechanism and procedure of proposed algorithm is explained. Section 5 analyzes the experimental result of proposed algorithm based on five benchmark complex functions and 0-1 knapsack problem. Finally, section 6 concludes the paper.

## 2. RELATED WORK

Genetic algorithms were formally introduced in the United States in the 1970s by John Holland at University of Michigan. To optimize complex numerical and combinatorial optimization problems genetic algorithm has been used in [16]-[19]. But the algorithm has the problems of premature convergence and it easily trap into local optimum. To overcome the problems a novel algorithm called Quantum- inspired Evolutionary Algorithm (QEA) has been introduced from the last decade in [4]-[9], [21]. But the algorithm still traps in local optimum when it solves the complex optimum problems because it uses only the information of the individual with optimum performance, but does not use the information of the individuals with the suboptimum performance of the population. Therefore, to solve complex numerical problem effectively a special algorithm called Real-coded Quantum Evolutionary Algorithm (RCQEA) has been proposed in [15].

A Particle Swarm Optimization (PSO) is a population based stochastic search technique which has also been applied to solve numerical and combinatorial optimization problem in [10], [22], [23]. The PSO uses both the values from global optimum and suboptimum solutions of the population in its evolutionary equation to keep balance of exploration and exploitation. In this paper a generalized approach, HRCQEA is proposed to solve both numerical and combinatorial optimization problem effectively. HRCQEA works with a mutation operator (SMM) designed based on PSO and Arithmetic Crossover (AC) which are applied on the population of real coded triploid chromosome. The experimental results show that the proposed method is superior to other techniques.

## 3. OVERVIEW OF SOME EVOLUTIONARY TECHNIQUES

### 3.1 Quantum Evolutionary Algorithm

In QEA, A quantum bit is defined as the smallest unit information in two- state computer [20] which is defined with a pair of numbers (α, β) as follows:

$$\begin{bmatrix} \alpha \\ \beta \end{bmatrix} \quad (1)$$

A Q-bit individual as a string of m Q-bits is defined as

$$\begin{bmatrix} \alpha_1 & \alpha_2 & \dots \alpha_m \\ \beta_1 & \beta_2 & \dots \beta_m \end{bmatrix} \quad (2)$$

where $\alpha_i$ and $\beta_i$ are the probability amplitudes of $i^{th}$ qubit and they satisfy the condition $|\alpha_i|^2 + |\beta_i|^2 = 1$, $i = 1, 2, 3\dots m$. $|\alpha_i|^2$ and $|\beta_i|^2$ gives the probability that the qubit will be found in '0' state and 1' state respectively. Q-bit representation has the advantage that it is able to represent a linear superposition of states [21]. The probability amplitudes of a qubit are updated by Q-gate which is a variation operator of QEA. After applying Q-gate, the qubit should satisfy the normalization condition $|\alpha'|^2 + |\beta'|^2 = 1$, where $|\alpha'|^2$ and $|\beta'|^2$ are the values of updated Q-bit. The following rotation gate is used as Q-gate:

$$U(\Delta\theta_i) = \begin{bmatrix} \cos(\Delta\theta_i) & -\sin(\Delta\theta_i) \\ \sin(\Delta\theta_i) & \cos(\Delta\theta_i) \end{bmatrix} \quad (3)$$

where $\Delta\theta_i$, $i = 1, 2, 3\dots, m$, is the rotation angle of a qubit towards the "0" state or "1" state depending on its sign. From the above description, it can be seen that solutions of QEA are represented directly or indirectly by binary string, which implies that QEA has some disadvantages such as the inconvenient process of coding and decoding. It does not adapt to the dimensions, the precision, and low search efficiency when applied for numerical optimization problems.

### 3.2 Real-coded Quantum Evolutionary Algorithm

RCQEA was proposed to solve complex numerical optimization problems. In RCQEA, a real-coded triploid chromosome is represented as follows [15]

$$\begin{pmatrix} x_1 \cdots x_i \cdots x_n \\ \alpha_1 \cdots \alpha_i \cdots \alpha_n \\ \beta_1 \cdots \beta_i \cdots \beta_n \end{pmatrix} \quad (4)$$

where $(x_i\ \alpha_i\ \beta_i)^T$, $i = 1, 2\dots n$ is the $i^{th}$ allele of real-coded triploid chromosome, $x_i$ is the real variable, a pair of probability amplitudes of one qubit is $(\alpha_i, \beta_i)^T$ which must satisfy the normalization condition $|\alpha_i|^2 + |\beta_i|^2 = 1$. Here, $n$ is the length of real-coded triploid chromosome.

### 3.3 Particle Swarm Optimization

PSO is a population-based, self-adaptive search optimization technique introduced by Kennedy and Eberhart in 1995 [10]. In PSO, each single solution is a "particle" in the search space. All of particles have fitness values which are evaluated by the fitness function to be optimized, and have velocities which direct the flying of the particles. The trajectory of each particle in the

174

search space is adjusted by dynamically altering the velocity of each particle, according to its own flying experience and the flying experience of the other particles in the search space. The position vector and the velocity vector of the $i^{th}$ particle in the D dimensional search space can be represented as $X_i = (X_{i1}, X_{i2}, \cdots, X_{iD})$ and $V_i = (V_{i1}, V_{i2}, \cdots, V_{iD})$, respectively [22]. At each generation, each particle is updated by following two 'best' values. The first one is the best previous location (the position giving the best fitness value) a particle has achieved so far. This value is called pBest. The pBest of the $i^{th}$ particle is represented as $P_i = (p_{i1}, p_{i2}, \cdots, p_{iD})$. Second one is the best position discovered by the whole population which is represented as $P_g = (P_{g1}, P_{g2}, \cdots, P_{gD})$, and called global best value. Then the new velocities and the position of the particle for the next fitness evaluation can be calculated using following two equations [22]:

$$V_{id} = V_{id} + C_1 r_1 (P_{id} - X_{id}) + C_2 r_2 (P_{gd} - X_{id}) \tag{5}$$

$$X_{id} = X_{id} + V_{id} \tag{6}$$

where $C_1$ and $C_2$ are acceleration coefficients and $r_1$ and $r_2$ are two separately generated uniformly distributed random numbers in the range [0, 1]. The first part of (5) represent the previous velocity provides the necessary momentum for particle to roam across the search space. The second part is the "cognitive" component, represents the personal thinking of each particle which encourages the particles to move their best positions found so far. The third component is the "social" component which helps to find the global best position found so far. From (5) it is seen that to find the next position of a particle, both the optimal $P_{gd}$ and suboptimal $P_{id}$ position is used. It means it does not easily fall in local optima.

## 4. PROPOSED ALGORITHM

QEA pilots the evaluation of the qubit chromosomes by applying the quantum gates. The algorithm often traps in local optimum when it solves the complex optimum problems because it uses only the information of the individual with optimum performance, but does not use the information of the individuals with the suboptimum performance. RCQEA also applies quantum rotation gate on the qubit chromosomes, but does not use any information of other individuals, even though it can solve complex optimization problems efficiently and effectively [15]. PSO finds the optimum solution through the communication of individuals of the swarm. Unlike QEA only considering the individuals with best performance, PSO not only uses the information of optimum individuals, but also the information of suboptimum individuals. So PSO has better global search ability than QEA. In addition, the evolutionary equation of PSO is simple that makes it easy and fast way. Combining the advantage of PSO and RCQEA, we proposed HRCQEA.

### 4.1 Mechanism of Proposed Algorithm

Let, S be the search space and $f : F \rightarrow R^n$ an objective function and $g_i : S \rightarrow R^n$, $i = 1, 2, \cdots, q$ set of functions (called constraints). The global optimization problem is then given as the task,

$$\text{Minimize } f(\mathbf{x}), \text{ such that } g_i(\mathbf{x}) < 0$$
$$\mathbf{x} = (x_1, \ldots, x_i, \ldots, x_n) \; \varepsilon \; S \tag{7}$$

where the subset $F$ is called the feasible region in S, $\mathbf{x} \; \varepsilon \; R^n$ defines the $n$ dimensional search space and each $x_i$ is bounded within $[x_{i,min}, x_{i,max}]$, where $i=1,2,3,\ldots, n$. Here the optimization problem has been specified as a minimization problem. This does not restrict the generality since every maximization problem can be specified as minimization problem using the following relation:

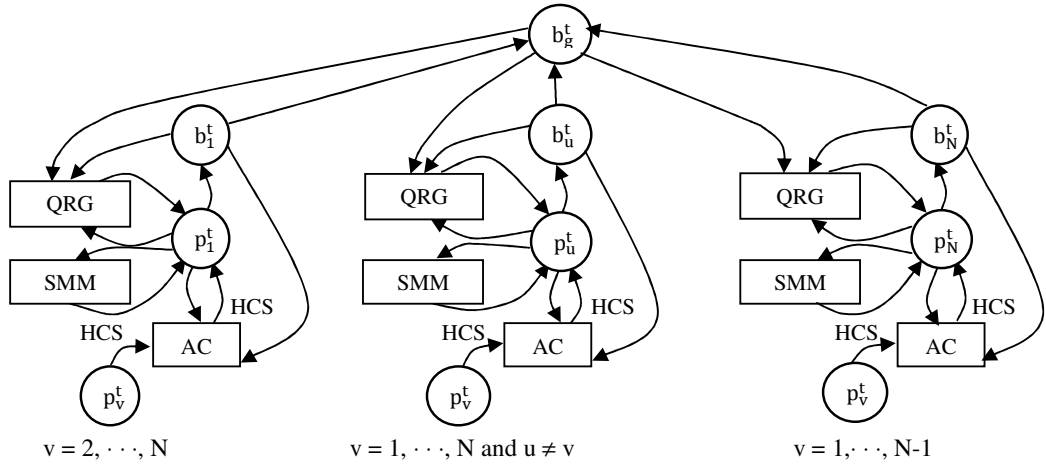

Figure 1. Overall Structure of HRCQEA

$$\max\{f(x)|x \in F\} = -\min\{-f(x)|x \in F\} \tag{8}$$

HRCQEA is proposed to optimize the complex numerical functions as well as constraint satisfaction problem such as knapsack problem and to accelerate the convergence speed of the evolutionary algorithm. Figure 1 describes the mechanism of HRCQEA. Where $p_u^t$ and $p_v^t$ are real-coded triploid chromosomes, u ≠ v, u, v = 1, …, N and they define $u^{th}$ and $v^{th}$ particle in the particle swarm. Here, each individual in the population is considered as a particle of the swarm. $b_u^t$ is the previous state with the best performance of the $u^{th}$ individual. $b_g^t$ is the global best solution which is the state with best performance so far in the neighborhood. To find the global optimum solution using HRCQEA, each particle $p_u^t$, u = 1, 2, · · ·N is represented by real-coded triploid chromosome which communicates with its own best position $b_u^t$ and global best position $b_g^t$ through QRG which uses the basic evolutionary equation of PSO. The position of $p_u^t$ is updated using the information gathered from the population by SMM. AC is used to expand the search space and HCS along with elitist selection is used for selecting elite.

### 4.1.1 Representation

Each particle or individual in HRCQEA is represented by the real-coded triploid chromosome as (4), where *n* is the length of the real-coded triploid chromosome i.e., the dimensions of function to be optimized, the vector $x = (x_1, \cdots, x_i, \cdots, x_n)$ represents the position of a particle that is a solution of optimization problem and $x_i$ is the real variable of the function to be optimized. $(\alpha_i, \beta_i)^T$ is a pair of probability amplitudes of the $i^{th}$ allele $(x_i, \alpha_i, \beta_i)^T$, $i = 1, 2 \cdots n$ of the triploid chromosome and satisfy the condition $|\alpha_i|^2+|\beta_i|^2 = 1$.

### 4.1.2 Mutation

In single-gene mutation, only one gene is mutated which is selected randomly from the individual chosen to mutate, but all other genes are fixed. Similarly, in multiple-gene mutation a number of genes are selected to mutate. In HRCQEA, single and multiple gene mutation are employed alternatively under a set of conditions. In Single-Multiple gene Mutation (SMM), Complementary Double Mutation Operator (CDMO) is used to update real variable of qubit(s) selected to mutate and QRG is used to update the pair of probability amplitudes of that qubit(s).

International journal of computer science & information Technology (IJCSIT) Vol.2, No.4, August 2010

In HRCQEA, a swarm of particles or a population of real-coded triploid chromosomes $P^t = \{p_1^t, \cdots, p_j^t, \cdots, p_N^t\}$ is maintained at generation t, where N is the size of population and $p_j^t$ is a particle or individual defined as (4) and $\{f_1^t, \cdots, f_j^t, \cdots, f_N^t\}$ describe the corresponding fitness values of individuals. Another population $B^t = \{b_1^t, \cdots, b_j^t, \cdots, b_N^t\}$ is used to hold the suboptimum solutions. For each particle $p_j^t$, an average rotation angle $\theta_j^t = \frac{1}{n}\sum_{i=1}^{n}\theta_{j,i}^t$ is maintained where $\theta_{j,i}^t$ is the rotation angle for the $i^{th}$ allele of $p_j^t$. In single gene mutation, choose the $i^{th}$ allele $(x_{j,i}^t\ \alpha_{j,i}^t\ \beta_{j,i}^t)^T$ randomly from all alleles of $p_j^t$ and update real variable $x_{j,i}^t$ as follows:

$$x_{j,i}^{t+1,k} = x_{j,i}^t + (x_{i,max} - x_{i,min}) * \Delta x * \xi \qquad (9)$$

Where $\Delta x$ and $\xi$ are defined as follows:

$$\Delta x = \sum_{g=1}^{\delta} r_g - \delta/2 \qquad (10)$$

Here $r_g$ is a random number in the range [0, 1]. For minimization problem,

$$\xi = \begin{cases} |\alpha_{j,i}^t|, & Fine\ search \\ |\beta_{j,i}^t|, & Coarse\ search \end{cases} \qquad (11)$$

For maximization problem,

$$\xi = \begin{cases} |\alpha_{j,i}^t|, & Coarse\ search \\ |\beta_{j,i}^t|, & Fine\ search \end{cases} \qquad (12)$$

Both of Fine and Coarse searches are applied repeatedly to mutate the qubits. It should be noted that the new real variable $x_{j,i}^{t+1,k}$ would exceed the limit of $x_{j,i}^t$ whether the value of $\xi$ is larger or the value of $x_{j,i}^t$ is close to the bound. To avoid making infeasible solution, new real variable $x_{j,i}^{t+1,k}$ is clipped as follows

$$x_{j,i}^{t+1,k} = \begin{cases} 2x_{i,max} - x_{j,i}^{t+1,k}, & x_{j,i}^{t+1,k} > x_{i,max} \\ 2x_{i,min} - x_{j,i}^{t+1,k}, & x_{j,i}^{t+1,k} < x_{i,min} \end{cases} \qquad (13)$$

Until $x_{j,i}^{t+1,k}$ lies in the feasible solution space, (13) has to be performed again and again.

If the new feasible solution $(x_{j,1}^t, \cdots, x_{j,i}^{t+1,k}, \cdots, x_{j,n}^t)$, which is derived from (9), (10),(11),(12) and (13), is better than the old feasible solution $(x_{j,1}^t, \cdots, x_{j,i}^t, \cdots, x_{j,n}^t)$, we call that the valid evolution is carried out, otherwise the invalid evolution is done. When the valid evolution occurs, the probability amplitudes $(\alpha_{j,i}^t\ \beta_{j,i}^t)^T$ are fixed, that is $\alpha_{j,i}^{t+1} = \alpha_{j,i}^t$, $\beta_{j,i}^{t+1} = \beta_{j,i}^t$. On the contrary once the invalid evaluation occurs, the probability amplitudes $(\alpha_{j,i}^t\ \beta_{j,i}^t)^T$ are updated by QRG as follows:

$$\begin{pmatrix} \alpha_{j,i}^{t+1} \\ \beta_{j,i}^{t+1} \end{pmatrix} = \begin{pmatrix} \cos(\theta_{j,i}^t) & -\sin(\theta_{j,i}^t) \\ \sin(\theta_{j,i}^t) & \cos(\theta_{j,i}^t) \end{pmatrix} \begin{pmatrix} \alpha_{j,i}^t \\ \beta_{j,i}^t \end{pmatrix} \qquad (14)$$

Here $\theta_{j,i}^t$ s the rotation angle and it is defined by the basic equation of the PSO as follows:

$$\theta_{j,i}^t = C_1(x_{j,i}^{b,t} - x_{j,i}^t) + C_2(x_{g,i}^t - x_{j,i}^t) \tag{15}$$

Where $C_1$ and $C_2$ are called the learning factor, $x_{j,i}^t$ is the $i^{th}$ real variable of the $p_j^t$  We have defined that $(x_{j,1}^{b,t} \cdots x_{j,i}^{b,t} \cdots x_{j,n}^{b,t})$ is the best position of the $j^{th}$ particle $p_j^t$ which is stored in $b_j^t$. $x_{g,i}^t$ is the $i^{th}$ real variable of the best performance so far in the neighborhood that is global best performance $b_g^t$ whose position is defined as $(x_{g,1}^t \cdots x_{g,i}^t \cdots x_{g,n}^t)$.

When the $i^{th}$ allele of $p_j^t$ causes continuously invalid evolution, a new approach will be used to update $\alpha_{j,i}^t$ and $\beta_{j,i}^t$ at larger scale apart from (14) and (15) so as to accelerate the convergence speed and achieve the aim of adaptive controlling of the evolutionary process of algorithms. Assume that $c_i$ is used to store the number of invalid evaluation occurred by $i^{th}$ allele. In every operation, whether "Fine search" or "Coarse search", if invalid evolution is occurred then real variable of $i^{th}$ allele will be hold, and the number of the continuously invalid evolution $c_i$ for that allele will be increased by 1. On the contrary, if valid evolution is occurred, real variable of the allele appointed will be replaced, and $c_i$ will be cleared. Depending on the value of $c_i$, $i = 1, 2, \cdots n$, alleles are selected from $p_j^t$ to update pair of probability amplitudes of these alleles. If $c_i$ for $i^{th}$ allele of $p_j^t$ is 0 then probability amplitudes $\alpha_{j,i}^t$ and $\beta_{j,i}^t$ are not changed. If the value of $c_i$ is less than or equal to a specified value $\lambda$ then (14) and (15) are used to update $\alpha_{j,i}^t$ and $\beta_{j,i}^t$, otherwise following equation is used to update them. For minimization problem,

$$\begin{cases} \alpha_{j,i}^{t+1} = \alpha_{j,i}^{t+1}/(fix(c_i/5) + 1) \\ \beta_{j,i}^{t+1} = \sqrt{1 - (\alpha_{j,i}^{t+1})^2} \end{cases} \tag{16}$$

For maximization problem,

$$\begin{cases} \beta_{j,i}^{t+1} = \beta_{j,i}^{t+1}/(fix(c_i/5) + 1) \\ \alpha_{j,i}^{t+1} = \sqrt{1 - (\beta_{j,i}^{t+1})^2} \end{cases} \tag{17}$$

Where $fix(\cdot)$ is a round function. From the above process, the allele $(x_{j,i}^t\ \alpha_{j,i}^t\ \beta_{j,i}^t)^T$ of $p_j^t$ is updated. For minimization problem, from (14), (15), and (16), it can be seen that with the increase of generation $t$, the value of $|\alpha_{j,i}^t|$ will decrease gradually and then the value of $\xi$ determined by (11) will reduce slowly if it is assigned to $|\alpha_{j,i}^t|$. So "Fine search" in the neighborhood of current solution is carried out. In reverse, the value of $|\beta_{j,i}^t|$ will increase gradually and then the value of $\xi$ will enhance slowly if it is assigned to $|\beta_{j,i}^t|$, that is " Coarse search" in the whole solution space is realized. Similarly for maximization problem, "Fine Search" and "Coarse Search" are realized subject to the value of $\xi$ determined by (12). "Fine search" in local search space and "Coarse search" in global search space make HRCQEA to treat the balance between exploration and exploitation, which is the origin of CDMO. "Fine search" and "Coarse search" are applied repeatedly for $m_1$ and $m_2$ times respectively for every individual in population and usually $m_1 > m_2$. In multiple-gene mutation, one or more qubits are selected randomly from $p_j^t$ and then (9) to (13) are used to update the real variables and pair of probability amplitudes is updated using (14) to (17) for all genes selected to mutate. The number of qubits to be selected for mutation is determined as follows:

$$n' = \left\lceil \frac{n}{4}\left(1 - \frac{t}{T_{max}+1}\right)\right\rceil \tag{18}$$

where, $T_{max}$ is the maximum number of generation used in the algorithm. The *ceil* function ensures that at least one qubit will be selected when $t$ become very close to $T_{max}$. The multiple-gene mutation is employed after a defined interval ($\kappa$) of generations if the value of,

$$\begin{cases} \theta_j^t < 0, & \text{Minimization problem} \\ \theta_j^t > 0, & \text{Maximization problem} \end{cases} \quad (19)$$

Where $\theta_j^t$ is the average rotation angle of $p_j^t$. From (15), it can be seen that for minimization problem $\theta_{j,i}^t$ that is $\theta_j^t$ will be negative if $p_j^t$ goes away from its own best position and the global best position of the population. Then multiple-gene mutation is applied to force the particle to move quickly towards the optimum solution. Similarly, for maximization problem $\theta_j^t$ will be positive for any invalid movement of $p_j^t$. Thus, the multiple-gene mutation improves the convergence speed of the algorithm.

### 4.1.3 Crossover

The crossover operator is designed based on the concept of AC [24]. In HRCQEA, crossover is performed after a fixed number of generations that is after a defined interval $\tau$ and $m$ times for each individual. The process of crossover operator can be explained as follows: For each individual $p_u^t$, $u = 1, 2, \cdots, N$, select randomly another individual $p_v^t$, $v = 1, 2, \cdots, N$, where u $\neq$ v. Then generate an individual,

$$P_{avg}^t = \begin{pmatrix} x_{avg,1}^t & \cdots & x_{avg,i}^t & \cdots & x_{avg,n}^t \\ \alpha_{avg,1}^t & \cdots & \alpha_{avg,i}^t & \cdots & \alpha_{avg,n}^t \\ \beta_{avg,1}^t & \cdots & \beta_{avg,i}^t & \cdots & \beta_{avg,n}^t \end{pmatrix} \quad (20)$$

Where,

$$\begin{cases} x_{avg,i}^t = (x_{u,i}^t + x_{v,i}^t)/2 \\ \alpha_{avg,i}^t = (\alpha_{u,i}^t + \alpha_{v,i}^t)/2 \\ \beta_{avg,i}^t = \sqrt{1 - (\alpha_{avg,i}^t)^2} \end{cases} \quad (21)$$

Now $P_{avg}^t$ and $b_u^t$ are considered as parents and two offspring $P_{d1}^t$ and $P_{d2}^t$ are generated from them as follows:

$$\begin{cases} x_{d1,i}^t = r_i \cdot x_{u,i}^{b,t} + (1 - r_i) \cdot x_{avg,i}^t \\ \alpha_{d1,i}^t = r_i \cdot \alpha_{u,i}^{b,t} + (1 - r_i) \cdot \alpha_{avg,i}^t \\ \beta_{d1,i}^t = \sqrt{1 - (\alpha_{d1,i}^t)^2} \end{cases} \quad (22)$$

$$\begin{cases} x_{d2,i}^t = (1 - r_i) \cdot x_{u,i}^{b,t} + r_i \cdot x_{avg,i}^t \\ \alpha_{d2,i}^t = (1 - r_i) \cdot \alpha_{u,i}^{b,t} + r_i \cdot \alpha_{avg,i}^t \\ \beta_{d2,i}^t = \sqrt{1 - (\alpha_{d2,i}^t)^2} \end{cases} \quad (23)$$

Where, $r_i$, $i = 1, 2, \cdots, n$ are random numbers, uniformly distributed in [0, 1]. $\left(x_{d1,i}^t \ \alpha_{d1,i}^t \ \beta_{d1,i}^t\right)^T$, $\left(x_{d2,i}^t \ \alpha_{d2,i}^t \ \beta_{d2,i}^t\right)^T$ and $\left(x_{u,i}^{b,t} \ \alpha_{u,i}^{b,t} \ \beta_{u,i}^{b,t}\right)^T$ are the $i^{th}$ allele of $P_{d1}^t$, $P_{d2}^t$ and $b_u^t$ respectively. In HRCQEA, AC would play an important role on preventing the individual in the early generations from being trapped in the local optima by expanding search space.

### 4.1.4 Hill-Climbing Selection (HCS) and Elitism

There are many HCS algorithms have been developed [25]. In HRCQEA, a very simple one of HCS algorithm has been used. If the offspring, which is formed from either mutation operator or AC, are superior to the parents, the parents are substituted for the offspring, otherwise the parents are saved. The above process is called "Hill-climbing" selection (HCS). It is obvious that HCS has advantages to guarantee the direction of search and accelerating the convergence speed. Along with HCS, elitist selection ensures that most fit members of each generation will be selected. Most of the GAs do not use pure elitism, but instead use a modified form where the single best or a few of the best individuals from each generation are copied into the next generation. In HRCQEA, elitism is used to store the own best positions of particles and the global best position of the swarm.

## 5. PERFORMANCE EVALUATION AND RESULTS ANALYSIS

### 5.1 Test Functions

To test the performance of HRCQEA five benchmark functions from [7] are used as test functions. These functions are described as follows-

*F1: Sphere Function*

$$\textbf{\textit{Minimize }} f(x) = \sum_{i=1}^{D} x_i^2 \tag{24}$$

where $-100 \leq x_i \leq 100$ and D=30. The global minimum value is 0.0 at $x = (0, 0 \cdots 0)$.

*F2: Rastrigin Function*

$$\textit{Minimize } f(x) = 10D + \sum_{i=1}^{D} (x_i^2 - 10\cos(2\pi x_i)) \tag{25}$$

where $-5.12 \leq x_i \leq 5.12$ and D=30. The global minimum value is 0.0 at $x = (0, 0 \cdots 0)$.

*F3: Ackley Function*

$$\textit{Minimize } f(x) = -20 \exp\left(-0.2\sqrt{\frac{1}{D}\sum_{i=1}^{D} x_i^2}\right) - \exp(\frac{1}{D}\sum_{i=1}^{D} \cos(2\pi x_i)) + 20 + e \tag{26}$$

where $-32 \leq x_i \leq 32$ and D=30. The global minimum value is 0.0 at $x = (0, 0 \cdots 0)$.

*F4: Schwefel Function*

$$\textit{Minimize } f(x) = 418.9829D - \sum_{i=1}^{D} x_i \sin(\sqrt{|x_i|}) \tag{27}$$

where $-500 \leq x_i \leq 500$ and D=30. The global minimum value is 0.0 at $x$ = (420.9687, 420.9687 ⋯ 420.9687).

*F5: Griewank Function*

$$\text{Minimize } f(x) = \frac{1}{4000} \sum_{i=1}^{D} x_i^2 - \prod_{i=1}^{D} \cos\left(\frac{x_i}{\sqrt{i}}\right) + 1 \tag{28}$$

where $-600 \leq x_i \leq 600$ and D = 30. The global minimum value is 0.0 at $x$ = (0, 0 ⋯ 0).

## 5.2 Results and Comparison for above five numerical problems

For performance comparison, five functions are solved by HRCQEA, RCQEA [15], QEA [7], and PSEQEA [22]. The maximum number of generation $T_{max}$ = 4000 as termination condition, population size N = 10 and run times 50 are used for all of four algorithms. For generalization, number of "Fine Search" and "Coarse Search" for RCQEA and HRCQEA are defined in terms of dimension of the function to be optimized as follows:

$$\begin{aligned} m1 &= 1.5 * D \\ m2 &= 0.5 * D \end{aligned} \tag{29}$$

Other parameters of four algorithms are initialized as follows:

a. **QEA**: The number of qubits for each variable that are used for F1, F2, F3, F4 and F5 are set to 28, 24, 26, 30 and 31 respectively.

b. **PSEQEA**: The number of qubits used for each variable for each functions is same as QEA. The value of c1 and c2 are set to $0.02\pi$.

c. **RCQEA**: The period of discrete crossover τ = 500. The number of continuous discrete crossover m = 10. The initial rotation angle $\theta_0 = 0.1\pi$, the scale parameter γ = 5.

d. **HRCQEA**: The period of multiple-gene mutation κ = 5. The value of τ = 500 and m = 10. The value of learning factors c1 and c2 are set to π, λ = 1 and δ = 12.

Table 1. Experimental Results of Four Algorithms on F1 ~ F5

| Functions | Algorithm | Best | Worst | Mean | σ |
|---|---|---|---|---|---|
| F1 | QEA | 2.211 | 11.35 | 5.28 | 1.90 |
|  | PESQEA | 0.0098 | 0.0520 | 0.0253 | 0.0095 |
|  | RCQEA | 7.2e-016 | 5.7e-015 | 2.0e-015 | 1.2e-015 |
|  | HRCQEA | 1.7e-140 | 3.3e-124 | 1.1e-125 | 5.4e-125 |
| F2 | QEA | 48.56 | 82.502 | 67.987 | 7.519 |
|  | PESQEA | 33.059 | 70.895 | 47.820 | 8.878 |
|  | RCQEA | 5.68e-14 | 3.97e-13 | 1.84e-13 | 7.75e-14 |
|  | HRCQEA | 0 | 0 | 0 | 0 |
| F3 | QEA | 3.829 | 8.027 | 6.241 | 0.9756 |
|  | PESQEA | 0.767 | 1.039 | 0.982 | 0.0588 |
|  | RCQEA | 3.9e-14 | 3.31e-7 | 2.48e-7 | 2.75e-8 |
|  | HRCQEA | 1.7e-007 | 1.7e-007 | 1.7e-007 | 3.5e-015 |
| F4 | QEA | 1785.7 | 3001.5 | 2381.2 | 280.42 |
|  | PESQEA | 80.14 | 3954.9 | 735.38 | 567.67 |
|  | RCQEA | 0.0004 | 236.88 | 35.532 | 59.82 |
|  | HRCQEA | 0.00039 | 0.00039 | 0.00039 | 0 |
| F5 | QEA | 2.437 | 9.738 | 5.664 | 1.690 |

| | PESQEA | 0.815 | 1.049 | 0.978 | 0.0591 |
| --- | --- | --- | --- | --- | --- |
| | RCQEA | 3.9e-14 | 4.7e-13 | 1.7e-13 | 1.17e-13 |
| | HRCQEA | 0 | 3.5e-15 | 3.5e-16 | 1.11e-15 |

Table 1 shows the experimental results of four algorithms on five functions each of dimensions D =30. It can be seen that the performance of QEA is the worst, which means that QEA is not suitable for numerical optimization problem. The PSEQEA is slightly superior to QEA. The performance of RCQEA is much better than QEA and PSEQEA, but HRCQEA has meaningfully better performance than RCQEA. Best, Worst and Mean denote the best fitness, worst fitness and mean fitness respectively over 50 runs. $\sigma$ denote standard deviation.

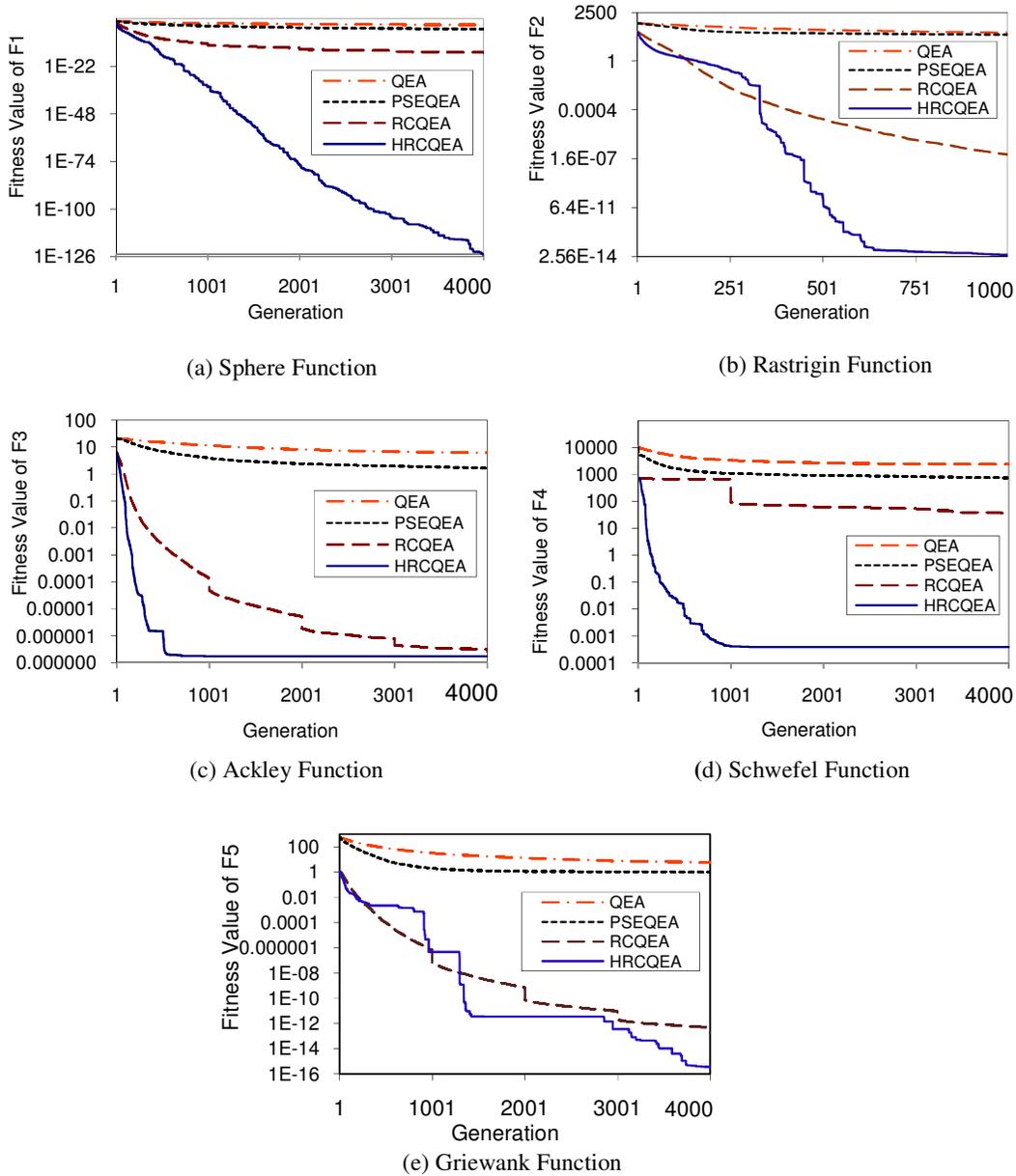

(a) Sphere Function

(b) Rastrigin Function

(c) Ackley Function

(d) Schwefel Function

(e) Griewank Function

Figure 2. Performance Comparison of four algorithms on functions F1 ~ F5

Figure 2 shows the progress of mean best fitness of functions F1 - F5 which is obtained by four algorithms with generations 4000. For rastrigin function (F2), fitness value has been shown only for 1000 generations because after 1000 generation HRCQEA gives the fitness value 0 which can't be plotted in the graph in logarithmic scale. It should be noted that RCQEA performs better than both QEA and PSEQEA in terms of search capability and convergence speed. The quality of the solutions and convergence speed of HRCQEA are superior to that of QEA, PSEQEA and RCQEA for all test functions. The complexity of an algorithm increases with increasing the dimensions of test functions. So it is necessary to observe the influence of the dimensions of test functions on the performance of HRCQEA. HRCQEA is also used to solve test functions by setting dimensions to 50 and 100. It is observed that HRCQEA can adjust itself with increasing the dimensions.

### 5.3 0-1 Knapsack Problem

In this subsection, the applicability of HRCQEA to the combinatorial optimization problem is described using knapsack problem. Let us consider that we have *n* objects or items and a knapsack or bag, Item *i* has a weight $w_i$, profit $p_i$ and the knapsack has capacity *C*. If an item *i* is placed into the knapsack then the profit of $p_i x_i$ is earned. The objective is to obtain a filling knapsack that maximizes the total profit earned. Since the knapsack capacity is *C*, we require the total weight of all chosen objects to be at most *C*. Formally the problem can be stated as maximize total profit,

$$f(x) = \sum_{i=1}^{n} p_i x_i \tag{30}$$

Subject to,

$$\sum_{i=1}^{n} w_i x_i \leq C \, , x_i = 0 \text{ or } 1 \tag{31}$$

If $x_i = 1$ then $i^{th}$ item is selected for the knapsack. The weights and profits are positive numbers. A feasible solution is any set $x = (x_1 \cdots x_n)$, satisfying (31). An optimal solution is a feasible solution for which (30) is maximized. If the knapsack is overfilled then the same repair method is used here as described in [21]. To solve the knapsack problem by HRCQEA, a binary version for each triploid chromosome is maintained. Let the binary version of $p_j^t$ is $Z_j^t = (z_{j,1}^t, \cdots z_{j,i}^t, \cdots , z_{j,n}^t)$ is constructed from real-coded triploid chromosome $p_j^t$ by the following make method:

*procedure make* ($p_j^t$ , $Z_j^t$)
**begin**
    $i \leftarrow 0$
    **while**( $i < n$ ) **do**
    **begin**
        **if** corresponding real variable of $x_{j,i}^t \geq 0.5$ **then**
            $z_{j,i}^t \leftarrow 1$
        **else**
            $z_{j,i}^t \leftarrow 0$
    $i \leftarrow i + 1$
    **end**
**end**

### 5.4. Results and Comparison for Knapsack Problem

For performance comparison of the HRCQEA, the knapsack problem is implemented by QEA and HRCQEA algorithm. For our test problem, data sets are generated as follows:

$$w_i = \text{uniformly random } [1 \cdots v] \text{ and } p_i = w_i + r$$

The average knapsack capacity is

$$C = \frac{1}{2}\sum_{i=1}^{n} w_i$$

Data are generated with the parameter settings: $r = 5$ and $v = 10$. Parameters for algorithms are initialized as follows:

a. **QEA**: The number of qubits for each individual is the number of items given.

b. **HRCQEA**: The value of learning factors $c_1$ and $c_2$ is set to $\pi$, value of $m_1$ and $m_2$ are set to 45 and 15 respectively. The period of multiple-gene mutation $\kappa = 1$ and all other parameters are set to same value as values set for numerical optimization problems.

**TABLE 2:** Experimental Results of Knapsack Problem

| No. of items | Algorithms | Best | Worst | Mean | $\sigma$ |
|---|---|---|---|---|---|
| 100 | QEA | 609 | 599 | 601.66 | 2.63888 |
| | HRCQEA | 614 | 604 | 609.26 | 1.52275 |
| 250 | QEA | 1522 | 1502 | 1508.34 | 5.08944 |
| | HRCQEA | 1567 | 1547 | 1556.46 | 4.6564 |

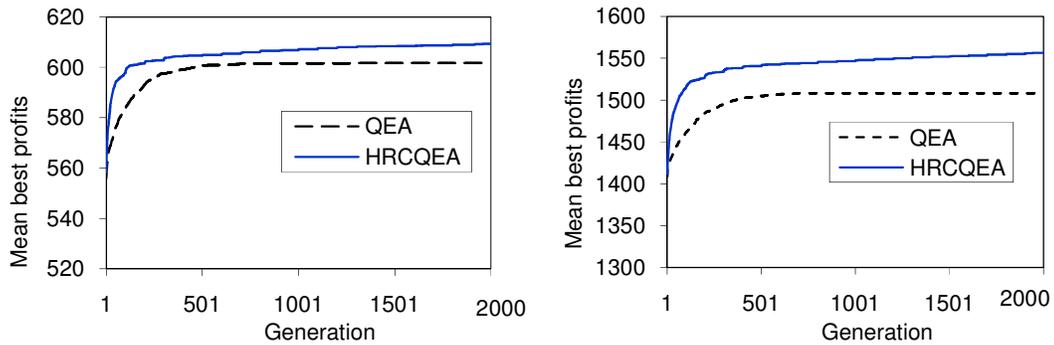

(a) Knapsack Problem for 100 Items      (b) Knapsack Problem for 250 Items

Figure 3. Performance Comparison of QEA and HRCQEA on Knapsack problem

For both algorithms, $T_{max} = 2000$ is used as termination condition and number of run = 50. Table 2 shows the experimental result of Knapsack problem obtained using QEA and HRCQEA for 100 and 250 items. Figure 3 shows the performance comparison of QEA and HRCQEA on 0-1 Knapsack problem for 100 and 250 items. It can be seen that the performance of HRCQEA is superior to QEA for 0-1 knapsack problem too.

### 5.5 Movement Analysis of the Population

All evolutionary algorithms always try to make the population to move towards the optimum solution. Convergence speed is one of the most important parameters to compare the performance of these algorithms. To compare the convergence speed of RCQEA and HRCQEA, average rotation angle for the global best particle was considered which indicates the average movement of the whole population.

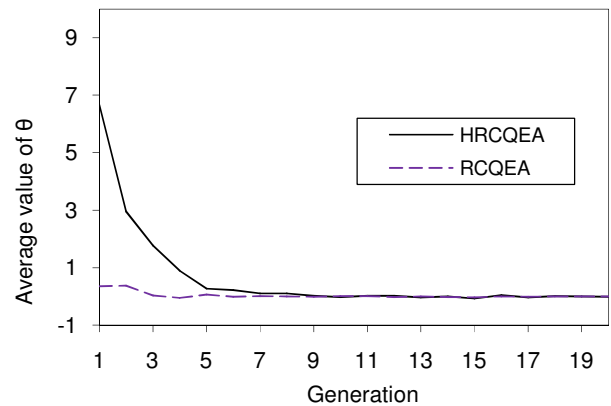

**Figure 4.** Average movement of population in RCQEA and HRCQEA

Figure 4 shows average rotation angle of the global best individual to solve sphere function with respect to generations. From (15), it can be seen that for minimization problem, high positive value of θ indicates the better movement of the population towards the optimum solution. The above figure shows very high value of θ for HRCQEA than RCQEA at early generations which implies that the HRCQEA has the highest convergence speed than RCQEA.

## 6. CONCLUSION

In this paper, a generalized hybrid algorithm named HRCQEA is proposed. The main idea of HRCQEA is to provide new techniques for variation operators based on Particle Swarm Optimization (PSO) and Arithmetic Crossover (AC). The evolutionary equation of PSO used in HRCQEA has more profound intelligent background which makes full use of information of the swarm. The crossover operator designed here also uses the information of suboptimum solutions of the swarm to expand the search space which helps to improve the convergence speed. Using the evolutionary equation of PSO, HRCQEA can adjust the rotation angle θ more reasonably. The average rotation angle maintained in this method helps the whole swarm to move towards the global best position very quickly. Thus, the HRCQEA can find the optimum solution faster. Some complex optimization problems and knapsack problem are used to test the performance of the proposed approach. The experimental results show that HRCQEA performs better than other algorithms in terms of global search capacity and convergence speed.

As the future work we will try apply the proposed approach to optimize the proportional gain and integral gain proportional-integral (PI) controller. This method also can be applied to solve multi-objective optimization problems.

International journal of computer science & information Technology (IJCSIT) Vol.2, No.4, August 2010

**Authors**

**Md. Amjad Hossain** received the B.Sc. Engg. degree (with honors) in Computer Science and Engineering (CSE) from Khulna University of Engineering & Technology(KUET), Khulna - 9203, Bangladesh, 2008. He is currently working as a Lecturer in Department of CSE, KUET. His current research interests include evolutionary computation and mobile computing.

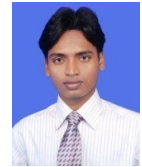

**Md. Kowsar Hossain** received the B.Sc. Engg. degree (with honors) in Computer Science and Engineering (CSE) from Khulna University of Engineering & Technology (KUET), Khulna - 9203, Bangladesh, in 2008. He is currently serving as a Lecturer in the Department of CSE of the same University. His current research interests include evolutionary computation and mobile computing.

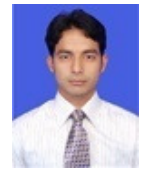

**M. M. A. Hashem** received the Bachelors degree in Electrical and Electronic Engineering from Khulna University of Engineering and Technology (KUET), Khulna, Bangladesh in 1988, Masters degree in Computer Science from Asian Institute of Technology (AIT), Bangkok, Thailand in 1993 and PhD degree in Artificial Intelligence Systems from Saga University, Japan in 1999. He is a Professor in the Dept. of Computer Science and Engineering, Khulna University of Engineering and Technology (KUET), Bangladesh.

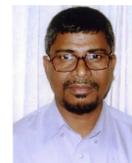

Currently, he is working as a Technical Support Team Consultant for Bangladesh Research and Education Network (BdREN) project of the University Grants Commission of Bangladesh. His research interest includes Evolutionary Computations, Intelligent Computer Networking, Wireless Networking, Soft-Computing, Evolutionary Cluster Computing etc. He has published more than 50 referred articles in international Journals/Conferences. He is a member of IEEE. He is a coauthor of a book titled Evolutionary Computations: New Algorithms and their Applications to Evolutionary Robots, Series: Studies in Fuzziness and Soft Computing, Vol. 147, Springer-Verlag, Berlin/New York, ISBN: 3-540-20901-8, (2004).